\documentclass[11pt]{article}
\usepackage[preprint]{acl}
\usepackage{times,latexsym}
\usepackage[T1]{fontenc}
\usepackage[utf8]{inputenc}
\usepackage{microtype}
\usepackage{inconsolata}
\usepackage{amsmath,amssymb}
\usepackage{graphicx}
\usepackage{booktabs}
\usepackage{array}
\usepackage{multirow}
\usepackage{algorithm}
\usepackage{algpseudocode}
\usepackage{xspace}
\usepackage{xcolor}
\usepackage{amsmath,amssymb,amsthm}

\newcommand{\lrx}{\ensuremath{\lambda_{\times}}\xspace}           
\newcommand{\lropt}{\ensuremath{\lambda_{\mathrm{ref\text{-}opt}}}\xspace}
\newcommand{\tauwin}{\ensuremath{\tau}\xspace}

\title{A Refresh Interval Does Not Fully Specify a Refresh Schedule: Learning-Rate-Dependent Placement in Kronecker-Preconditioned Language Model Training}

\author{Xinyu Wang\textsuperscript{1} \quad Zhiwei Zhai\textsuperscript{2} \quad Yu Gu\textsuperscript{1} \\
\textbf{Zijun Yu\textsuperscript{1} \quad Sicheng Lyu\textsuperscript{1} \quad Jingrui Tian\textsuperscript{1}} \\
\textbf{Lei Ding\textsuperscript{3} \quad Ziyang Song\textsuperscript{4} \quad Xiao-Wen Chang\textsuperscript{1}} \\[0.4em]
\textnormal{\small \textsuperscript{1}McGill University \quad \textsuperscript{2}The Chinese University of Hong Kong} \\
\textnormal{\small \textsuperscript{3}University of Manitoba \quad \textsuperscript{4}Ohio University}}

\begin{document}
\maketitle

\begin{abstract}
Kronecker-preconditioned optimizers recompute a cached curvature
eigendecomposition only periodically, and implementations and adaptive
methods control this through the refresh \emph{frequency}. We show that frequency leaves part of a schedule free: at a fixed number of
refreshes, \emph{when} they occur changes held-out loss at the same
compute. Training
GPT-style models with the Mousse optimizer across model sizes, we move the middle
refreshes of a four-refresh schedule while fixing the first and last. On FineWeb the two placements exchanged rank as the learning rate rose
(later placement was better at low rates, earlier placement at high ones),
so a single-learning-rate comparison can report either ranking. The
reversal held under controls including the published optimizer internals,
and placement still mattered at the published refresh density. The
learning rate at which the preference flips fell with model size,
consistent with a held-out smaller model, and shifted with training length.
A spectral-width staleness score predicted the same ranking at every
learning rate and missed the reversal.
Refresh schedules should be reported in full and compared across the
learning rates at which they will be used.

\end{abstract}

\section{Introduction}
\label{sec:intro}

\begin{figure}[t]
\centering
\includegraphics[width=0.99\columnwidth]{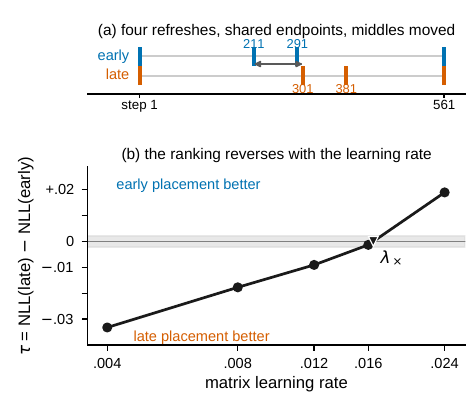}
\caption{(a)~The two schedules. Both perform four refreshes and share their
first and last refresh step, and only the two middle refreshes differ.
(b)~Seed-mean difference $\tauwin$ in endpoint held-out negative
log-likelihood (NLL), later schedule minus earlier, against the matrix
learning rate (FineWeb, 350M). Positive values mean earlier placement is
better. \lrx marks the interpolated sign change, and
the grey band is the tie tolerance.}
\label{fig:teaser}
\end{figure}

Kronecker-preconditioned optimizers
\citep{gupta2018shampoo,anil2020scalable,vyas2025soap,zhang2026mousse}
are increasingly competitive for neural-network and language-model training
\citep{kasimbeg2025algoperf}, but their curvature factorization is
expensive, so practical implementations compute it only intermittently and
reuse the cached result in between. The standard control on this cost is
the refresh interval, and recent methods adapt that interval to a staleness
signal \citep{eschenhagen2025purifying,nam2026foam}. Both treat a schedule
as determined by how often refreshes occur.

A refresh frequency fixes how many factorizations are computed and leaves
open the steps at which they run. The same number of refreshes can be placed
differently over training, and each placement decides which cached
curvature is applied to which updates. Moving a refresh costs the same compute, because the same
eigendecompositions run at different steps. We isolate this choice by holding the refresh count and the first and
last refresh steps fixed and moving only the middle refreshes. On FineWeb the same two schedules exchanged rank as the learning rate
rose. Later placement gave lower loss at small learning rates and earlier
placement at large ones (Figure~\ref{fig:teaser}). A comparison made at a
single learning rate therefore reports whichever ranking that rate
selects. This is the confound documented for comparisons between
optimizers \citep{choi2019empirical,sivaprasad2020optimizer}, here inside
one optimizer's schedule space.

\paragraph{Experimental design.}
All experiments use Mousse \citep{zhang2026mousse}, a curvature-augmented
Muon optimizer whose factored-curvature eigensystem is recomputed only at
refresh steps, on GPT-style models of 64M to 929M parameters trained in a
short-horizon regime on six single-domain corpora, with FineWeb as the
primary corpus (\S\ref{sec:experiments}).

\paragraph{Findings.}
\textbf{Placement preference reverses with the learning rate}
(\S\ref{sec:reversal}). Across intermediate placements, the lowest-loss
placement moved earlier as the learning rate rose.
\textbf{The reversal survives controls, and a uniform schedule beats both.}
It held under controls for warmup, weight decay, the auxiliary learning
rate and the published Mousse internals, and placement still changed loss
at the published refresh density. An equal-budget uniform schedule beat
both mirrored schedules on most corpora, so the mirrored pair serves as a
probe of how schedules are compared, and the uniform schedule is the
better choice among the three.
\textbf{The boundary moves with scale and horizon} (\S\ref{sec:transfer}).
It fell with model size, a held-out smaller model landed inside the band
fixed in advance, and it fell faster than the loss-optimal learning rate.
Longer training also shifted it.
\textbf{A spectral-width staleness score predicts the same ranking at every
learning rate} (\S\ref{sec:staleness}). Most spectral-width change happened
before the schedules diverge, so the score tracks early curvature motion
and misses the reversal. A schedule comparison therefore needs the full schedule, several learning
rates and an equal-budget uniform baseline.

\section{Related Work}
\label{sec:related}

\paragraph{Kronecker-factored preconditioning and its amortization.}
Kronecker-factored approximate curvature (K-FAC)
\citep{martens2015kfac}, Shampoo \citep{gupta2018shampoo} and their
large-scale refinements \citep{anil2020scalable,morwani2025newperspective,vyas2025soap}
precondition matrix-shaped parameters with factored curvature whose
eigendecomposition or inverse root dominates per-step cost, so
implementations recompute it intermittently. Eigenvalue-corrected K-FAC
(EKFAC) \citep{george2018ekfac} and SOAP, which runs Adam in the
eigenbasis of Shampoo's preconditioner \citep{vyas2025soap}, keep a diagonal rescaling current inside a
fixed eigenbasis between rebuilds. Standard Muon
\citep{jordan2024muon,bernstein2024old,liu2025muon} orthogonalizes the momentum directly, without a cached curvature
factorization. Mousse \citep{zhang2026mousse}
adds Kronecker-factored curvature to a Muon-style update and with it that
decision (\S\ref{sec:optimizer}). Appendix~\ref{app:survey} surveys the
refresh controls these implementations expose.

\paragraph{Deciding when to refresh.}
The closest work adapts the refresh frequency to an error signal. Purifying
Shampoo triggers an eigenbasis update from a warm-started QR termination
bound \citep{eschenhagen2025purifying}, and FOAM couples the
eigendecomposition frequency and damping to an estimated operator error
\citep{nam2026foam}. Both tie timing to cost: a refresh occurs when the
error is large enough, so changing when refreshes occur also changes how
many occur. Purifying Shampoo further finds that eigenbasis accuracy matters
more early in training than late, and leaves open how approximation error
affects convergence. \S\ref{sec:staleness} tests whether a scalar error
signal of this kind predicts the effect of placement. Non-uniform timing is already expressible in practice:
Distributed Shampoo exposes a warm-start step alongside its update interval
\citep{shi2023shampoo}.

\paragraph{Evidence on refresh frequency.}
SOAP reports that recomputing Shampoo's eigendecomposition less often
degrades language-model pretraining, increasingly so as the interval grows,
and that SOAP degrades more slowly \citep{vyas2025soap}. Such studies vary how often the factorization is recomputed, while our
experiments vary placement at a fixed count (\S\ref{sec:reversal}). Near
the preference boundary the
placement effect falls within the tie tolerance, so a comparison made there
would report little difference between schedules that separate clearly at
other learning rates.

\paragraph{Rankings that depend on how the comparison is run.}
Optimizer rankings are known to change with the tuning protocol:
comparisons at a single hyperparameter point, or under unequal tuning
budgets, can reverse conclusions
\citep{choi2019empirical,sivaprasad2020optimizer,schmidt2021crowded,dahl2023benchmarking}.
Recent language-model studies sweep the learning rate for every
method \citep{zhao2025deconstructing,semenov2025benchmarking} and treat
scale and data-to-model ratio as axes of the comparison
\citep{wen2025fantastic}, and \citet{wen2025fantastic} also observe that the
ranking of two optimizers can flip during learning-rate decay. Optimal
learning rates themselves shift with width \citep{yang2022mup} and token
horizon \citep{bjorck2024tokenhorizon}, the axes against which
\S\ref{sec:transfer} measures the boundary. We find the same kind of
reversal between two equal-cost schedules of one optimizer.

\section{Refresh Placement at Fixed Cost}
\label{sec:setup}

\subsection{Refresh schedules}
\label{sec:optimizer}
Consider a run of $H$ optimization steps. A refresh schedule with refresh
budget $K$ (the number of refreshes) is an ordered
set
\begin{equation}
S=(s_1,\ldots,s_K),\qquad 1=s_1<\cdots<s_K\le H. \notag
\end{equation}
A refresh is performed before the update at its scheduled step. For update
$t$, the most recent refresh and the age of its cached factorization are
\begin{equation}
r_S(t)=\max\{s\in S: s\le t\},\;\; a_S(t)=t-r_S(t).
\label{eq:lastrefresh}
\end{equation}
The optimizer we study, Mousse \citep{zhang2026mousse}, updates its momentum
and factored curvature statistics at every step but recomputes the
eigendecomposition of the curvature factors only at the steps in $S$,
applying the cached eigensystem to every update in between. The exact update
is given in App.~\ref{app:impl}. Changing $S$ therefore changes which
curvature snapshot is applied during each part of training, while the
number of eigendecompositions stays the same.

Our primary comparison uses $H=591$ steps with
\begin{equation}
\begin{aligned}
S_E&=(1,211,291,561),\\
S_L&=(1,301,381,561).
\end{aligned}\notag
\end{equation}
Here $S_E$ places the middle refreshes earlier and $S_L$ later. Both
schedules perform four refreshes and share a terminal cache
age of $591-561=30$ steps, and their middle refreshes lie within
the constant-learning-rate phase of training. Between steps 211 and 290,
for example, $S_E$ applies the cache computed at step 211 while $S_L$ still
applies the cache from step 1. From step 381 to step 560, $S_L$ applies its
step-381 cache while $S_E$ applies the one from step 291.

The pair also differs in total cache age. The sum $\sum_t a_S(t)$
depends on the refresh gaps and is minimized by even spacing, so the pair
differs in total staleness as well as in timing (App.~\ref{app:math}). Two designs separate these: the
placement landscape of \S\ref{sec:reversal} moves the middle pair while
keeping its separation fixed, and an equal-budget uniform schedule, which
minimizes cumulative age, serves as a reference throughout.

Placement also changes the contents of later caches. An earlier refresh
alters subsequent parameters, gradients, momentum and curvature
statistics, so even the shared final refresh can produce different cached
eigensystems on the two trajectories.

\subsection{Placement contrast and preference boundary}
\label{sec:estimand}
Let $\ell(S,\eta;\xi)$ denote endpoint held-out NLL at matrix learning rate
$\eta$ and seed $\xi$, with corpus, model size and horizon fixed. Schedules
are compared within matched seeds:
\begin{equation}
\begin{aligned}
\tauwin(\eta;\xi)&=\ell(S_L,\eta;\xi)-\ell(S_E,\eta;\xi),\\
\bar\tau&(\eta)=\frac{1}{n}\sum_{i=1}^{n}\tauwin(\eta;\xi_i),
\end{aligned}
\label{eq:tau}
\end{equation}
so positive \tauwin favors earlier placement. A contrast inside
$[-\delta,\delta]$ is treated as a tie, and a cell is seed-consistent when
every seed receives the same label (earlier better, later better, or tie).
The choice of $\delta$ is given in \S\ref{sec:experiments}. We report
individual seed contrasts with each comparison so that effect size and
run-to-run variation can be read separately (App.~\ref{app:surface}).

We call the learning rate at an observed sign change the \emph{preference
boundary} \lrx, or the crossing. A pair of adjacent grid points whose
seed-mean contrasts have opposite signs is a \emph{bracket}. We estimate
\lrx by interpolating the seed-mean contrast in log learning rate inside
the last bracket of the grid, staying inside the tested grid (Eq.~\ref{eq:crossing},
App.~\ref{app:math}). The seed contrasts on either side of \lrx show
whether the ranking changed consistently across runs.

\section{Experiments}
\label{sec:exps}

All experiments in this section share the setup below.

\subsection{Experimental setup}
\label{sec:experiments}

\paragraph{Models, data and training budget.}
We trained GPT-style decoder models \citep{radford2019gpt2} at five sizes from 64M to 929M
parameters on six single-domain corpora
(FineWeb \citep{penedo2024fineweb}, and the GitHub, Pile-CC,
PubMed~Central, arXiv and Books3 subsets of the Pile \citep{gao2020pile})
and on three FineWeb/GitHub mixtures. FineWeb was the primary corpus.
Unless stated otherwise, runs used 65{,}536 tokens per step for 591 steps
under a warmup--stable--decay (WSD) learning-rate schedule
\citep{hu2024minicpm,hagele2024wsd} (App.~\ref{app:impl}).
This is a deliberately short regime, between $0.04$ and $0.6$ tokens per
parameter and far below compute-optimal training
\citep{hoffmann2022chinchilla}. It made dense learning-rate and seed grids
affordable at five scales, and \S\ref{sec:transfer} varies the horizon
directly.

\paragraph{Optimizer.}
Matrix parameters used Mousse with the schedules of \S\ref{sec:setup}, and non-matrix parameters used AdamW \citep{loshchilov2019adamw}. Our implementation follows the published
update but differs in three internals: AdamW rather than Lion \citep{chen2023lion} for non-matrix parameters, a nonnegative eigenvalue correction in place of
additive damping, and a $\sqrt{m/n}$ learning-rate adjustment for an $m\times n$ weight matrix
(App.~\ref{app:impl}). A control in \S\ref{sec:reversal} restores all three
jointly.

\paragraph{Refresh budget.}
Four refreshes is sparse next to Mousse's published interval of ten steps,
although Shampoo-family deployments report intervals from tens to hundreds
of steps \citep{anil2020scalable,shi2023shampoo}. A sparse budget leaves
long gaps whose placement can be moved over a wide range.
\S\ref{sec:reversal} also tests larger budgets and the published refresh
density.

\paragraph{Learning rates and seeds.}
The corpus surface, the grid of corpora and matrix learning rates (LRs) at
350M, used five learning rates with two fresh seeds per cell. Controls used
three seeds. Independent cohorts, which repeat a model size with fresh
seeds, used four. Grids were fixed per experiment and are listed with each experiment
(App.~\ref{app:registration}). A corpus's tuned learning rate is the grid
value that minimizes endpoint NLL for the better of the two mirrored
schedules. The uniform reference schedule is $(1,188,374,561)$.

\paragraph{Evaluation and tie tolerance.}
Endpoint NLL was measured on a frozen held-out stream from a shard
disjoint from training. The tie tolerance
$\delta=0.002$ NLL was registered before the first surface experiment and
carried unchanged. It is about one unit of seed-to-seed variation. A run
counts as unstable when its endpoint NLL exceeds the best endpoint of its
condition by more than a fixed stability margin (App.~\ref{app:impl}). The conclusions rest on
contrasts several times larger (App.~\ref{app:surface}). Each experiment's protocol and
analyzer were fixed before its data existed (App.~\ref{app:registration}).

\subsection{Learning Rate Reverses Placement Preference}
\label{sec:reversal}

\begin{figure*}[t]
\centering
\includegraphics[width=1\textwidth]{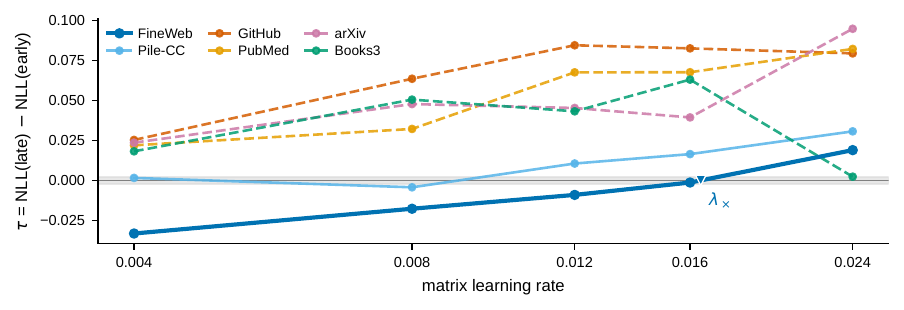}
\caption{The placement contrast $\tauwin(\mathrm{corpus},\eta)$ at GPT-350M
(positive $=$ early better). FineWeb is drawn heavier, the grey band is the
$\pm0.002$ tie tolerance, and the marker is the interpolated crossing.
Crossing corpora are solid, non-crossing dashed.}
\label{fig:surface}
\end{figure*}

Figure~\ref{fig:surface} compares
$S_E$ and $S_L$ at five learning rates on each of the six corpora at
GPT-350M, with two fresh seeds per cell. Every cell was seed-consistent
under the screening tolerance (Figure~\ref{fig:seedconsistency}). On
FineWeb, later placement gave lower loss at low learning rates. Its
advantage shrank as the learning rate rose, passed through a tie, and
reversed at the highest rate. The other corpora behaved differently. Pile-CC ended with earlier
placement ahead after a non-monotone path, GitHub, PubMed~Central and arXiv
favored earlier placement at every tested rate, and Books3 favored earlier
placement until its runs became unstable at the highest rate. The
three-seed controls below reproduce the FineWeb reversal.

\begin{table}[t]
\centering\footnotesize
\setlength{\tabcolsep}{7pt}
\caption{Placement effect at each corpus's tuned learning rate beside the
largest effect on the surface (350M, 591 steps).}
\label{tab:tuned}
\begin{tabular}{@{}lrrl@{}}
\toprule
corpus & tuned LR & $|\tauwin|$ tuned & surface max \\
\midrule
FineWeb & $.012$ & $.009$ & $.033$ (at $.004$) \\
Pile-CC & $.012$ & $.011$ & $.031$ (at $.024$) \\
GitHub & $.012$ & $.084$ & $.084$ (at $.012$) \\
PubMed & $.012$ & $.068$ & $.082$ (at $.024$) \\
arXiv & $.016$ & $.039$ & $.095$ (at $.024$) \\
Books3 & $.012$ & $.043$ & $.063$ (at $.016$) \\
\bottomrule
\end{tabular}

\end{table}

Table~\ref{tab:tuned} reports the effect at each corpus's tuned learning
rate, the operating point a practitioner would use. The effect remained beyond the tie tolerance on every corpus at
these points. It was smallest on FineWeb and largest on GitHub, and on
most corpora it was smaller than the largest value on the surface. For
scale, the tuned FineWeb effect is about $15\%$ of the loss reduction from
doubling the model from 350M to 695M parameters, obtained here at an unchanged number of eigendecompositions.

\paragraph{Comparison with uniform placement.}
At the
learning rates tuned for the mirrored pair, an equal-budget uniform
schedule outperformed the better mirrored schedule on most corpora
(App.~\ref{app:dominance}).
The two comparisons answer different questions. The mirrored pair measures
how a fixed change in timing interacts with the learning rate, and the
uniform schedule measures how far both placements sit from even spacing.
Among the schedules tested, even spacing is therefore a strong default at a
fixed budget, and the mirrored pair shows how much an uneven choice can shift
the ranking between learning rates.

\begin{figure}[t]
\centering
\includegraphics[width=0.98\columnwidth]{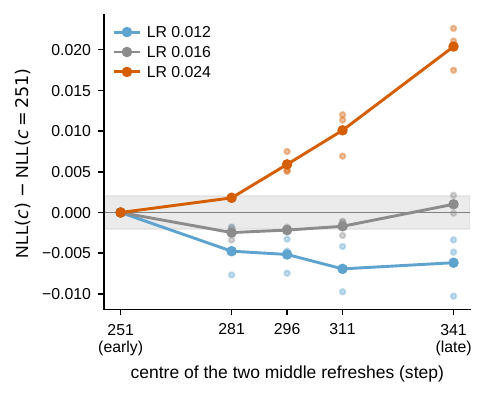}
\caption{Endpoint NLL relative to the earliest centre $c{=}251$ for five placements
of the middle pair at the same budget and endpoints (FineWeb, 350M). Faint
points are individual seeds, and the grey band is the tie tolerance.}
\label{fig:landscape}
\end{figure}

\paragraph{The placement landscape.}
To check whether the reversal extends beyond the two endpoints, we moved
the middle pair through
\begin{equation}
\begin{aligned}
S_c&=(1,\,c-40,\,c+40,\,561),\\
c&\in\{251,281,296,311,341\},
\end{aligned}
\label{eq:landscape}
\end{equation}
which keeps the refresh count, the endpoints and the middle-pair separation
fixed and has $S_E$ and $S_L$ as its two extremes. The whole family tilted
with the learning rate (Figure~\ref{fig:landscape}). At the lowest of the
three rates the earliest centre was the worst placement, at the highest it
was the best, and in between the landscape was nearly flat. At the highest
rate the end-to-end tilt was an order of magnitude larger than the tie
tolerance, while the best sampled centre and its neighbours differed by
about the tolerance. All three seeds had the same sign at the two end
columns.

\paragraph{Controls.}
Each control in Table~\ref{tab:controls} changed one part of the setup and
re-measured the FineWeb bracket. Adding a refresh at the end of warmup to
both schedules left the crossing nearly unchanged. Raising the budget while
keeping the first and largest gaps also left it nearly unchanged, so the reversal holds with more than two middle refreshes. Scaling
the AdamW rate with the matrix rate and varying weight decay both kept the
bracket. Restoring the published internals within our training recipe
moved the crossing to a higher learning rate. On FineWeb/GitHub mixtures
earlier placement won at every tested learning rate, so the late preference
at low rates appeared only on pure FineWeb.

\begin{table}[t]
\centering\footnotesize
\setlength{\tabcolsep}{4pt}\renewcommand{\arraystretch}{1.1}
\caption{Controls on the FineWeb reversal (350M, primary pair unless
stated). A check mark means a seed-consistent sign change, a cross means
none, and \lrx is the interpolated crossing. Details are in
App.~\ref{app:surface}.}
\label{tab:controls}
\begin{tabular}{@{}lcc@{}}
\toprule
condition & reversal & \lrx \\
\midrule
primary pair (reference) & \checkmark & $.0166$ \\
\addlinespace[3pt]
\multicolumn{3}{@{}l}{\emph{schedule construction}} \\
\quad shared refresh added at warmup end & \checkmark & $.0164$ \\
\quad budget 8, first and largest gaps kept & \checkmark & $.0166$ \\
\quad budget 16, first and largest gaps kept & \checkmark & $.0171$ \\
\quad budget 59, every gap at most 15 steps & ---$^{a}$ & --- \\
\addlinespace[3pt]
\multicolumn{3}{@{}l}{\emph{auxiliary optimizer}} \\
\quad AdamW rate scaled with the matrix rate & \checkmark & ---$^{b}$ \\
\quad weight decay varied & \checkmark & ---$^{b}$ \\
\addlinespace[3pt]
\multicolumn{3}{@{}l}{\emph{implementation}} \\
\quad published Mousse internals & \checkmark & $.032$ \\
\addlinespace[3pt]
\multicolumn{3}{@{}l}{\emph{data}} \\
\quad FineWeb/GitHub mixtures & $\times$ & --- \\
\bottomrule
\multicolumn{3}{@{}p{0.96\columnwidth}@{}}{\scriptsize $^{a}$Placement
changed loss, but the two highest-rate cells were seed-inconsistent
(Table~\ref{tab:dense}). $^{b}$Sign change kept in the same grid interval
as the reference. Per-seed values in Table~\ref{tab:ratio}.}
\end{tabular}

\end{table}

\begin{figure}[t]
\centering
\includegraphics[width=0.98\columnwidth]{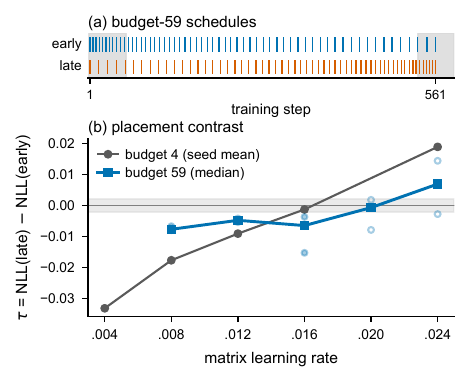}
\caption{(a)~The budget-59 schedules as tick marks (grey bands are WSD
warmup and warmdown). (b)~\tauwin against the matrix learning rate on
FineWeb at 350M for the budget-59 pair (median and individual seeds) and
the primary budget-4 pair (seed mean). Filled seed points mark
seed-consistent cells, open points the others, and the grey band is the tie
tolerance. Per-seed values are in Table~\ref{tab:dense}.}
\label{fig:density}
\end{figure}

\paragraph{Published refresh density.}
The larger budgets above keep the long gaps of the sparse pair. A budget-59
pair at the published refresh density instead keeps every gap short, with
interval lengths running in opposite directions
(Figure~\ref{fig:density}a). Placement still changed endpoint loss by
several times the tie tolerance, and every seed-consistent cell favored
later placement, as the sparse pair does at lower learning rates
(Figure~\ref{fig:density}b). At low rates the dense contrast was smaller
than the sparse one, and its seed spread grew at high rates. The two
highest-rate cells were seed-inconsistent, so at this density the data
show a placement effect and leave the reversal unresolved.

\subsection{The Boundary Varies with Scale and Horizon}
\label{sec:transfer}

\begin{figure*}[t]
\centering
\includegraphics[width=0.99\textwidth]{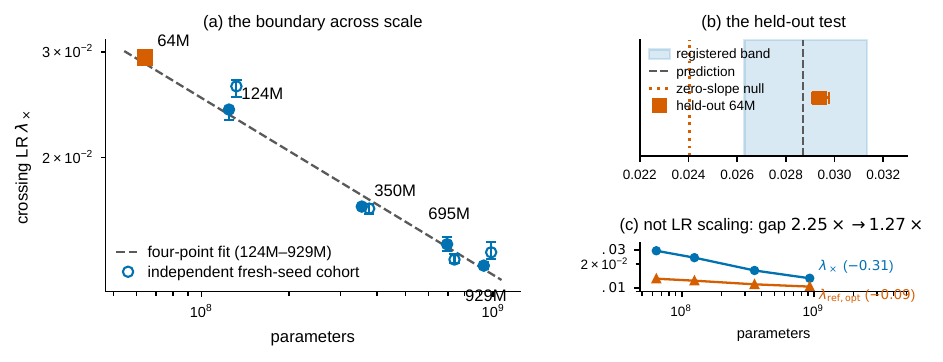}
\caption{(a)~\lrx against parameter count, one experiment per scale. The
dashed line is the fit to the four larger scales, and the square is the
held-out 64M measurement. Filled markers are the main estimates, and open
markers are independent fresh-seed cohorts (App.~\ref{app:ladder}).
(b)~The held-out test: the preselected tolerance band, the zero-slope
benchmark and the 64M measurement.
(c)~\lrx against the uniform-schedule optimal learning rate \lropt.}
\label{fig:ladder}
\end{figure*}

We measured \lrx for the same schedule pair on
FineWeb at five model sizes, each in a separate experiment with the
horizon of \S\ref{sec:experiments}. The boundary fell at every step up the
ladder (Figure~\ref{fig:ladder}a). To test the trend before seeing the
result, we fitted the four larger sizes and fixed a tolerance band around
the extrapolated 64M boundary before running 64M. The measured boundary
fell inside the band, much closer in log distance to the prediction than
to a benchmark in which the boundary stays constant across sizes (Figure~\ref{fig:ladder}b). The band is a preselected acceptance
tolerance, and a fit through the independent cohorts alone would have
placed the 64M boundary outside it (App.~\ref{app:ladder}).

Repeating three sizes with fresh four-seed cohorts tested how precisely
each size is located. Every cohort showed the same sign pattern across
learning rates, yet each placed its crossing outside the interval of the
main estimate, sometimes above it and sometimes below. The spread between
cohorts is thus wider than the spread between seeds, so the ladder fixes
the direction of the trend more firmly than any single per-size value.

\paragraph{Preference boundary and optimal learning rate.}
A downward boundary could simply track the loss-optimal learning rate,
which also falls with size. We compared the two directly at four scales,
using the learning rate that minimizes loss under the equal-budget uniform
schedule, \lropt. A pre-specified test checked whether the two log--log slopes agree within
an equivalence band. The slope difference and its whole seed-bootstrap
interval lay outside that band (App.~\ref{app:ladder}). The boundary sat above \lropt
at every measured scale, and the gap narrowed while staying open
(Figure~\ref{fig:ladder}c). Over 64M to 929M, the preference boundary and
the loss-optimal learning rate are therefore distinct quantities.
The boundary also differs from the stability edge, the lowest unstable
learning rate above the uniform-schedule optimum. Across scales the
boundary fell faster than the edge, and the ratio between them shrank
(App.~\ref{app:ladder}), so the crossing lies below the stability edge at
every measured scale.

\paragraph{Training horizon.}
All model sizes above share one token budget, so size and tokens per
parameter vary together along the ladder. At 350M, with the refresh
windows scaled in proportion, the reversal remained at twice and four times
the original horizon. At eight times the horizon every tested cell favored later placement or
tied, which leaves two readings open: the boundary moved above
the grid, or the contrast flattened (App.~\ref{app:ladder}). The smaller 64M model lost the effect sooner. By four times the original horizon, every contrast away from the stability
edge was within the tolerance. At matched
tokens per parameter, 350M crossed while 64M and 124M favored earlier
placement at every tested learning rate, so model size affects the
contrast beyond its effect on tokens per parameter (App.~\ref{app:ladder}).

\subsection{Spectral-Width Drift Predicts the Same Ranking at Every Learning Rate}
\label{sec:staleness}

\begin{figure*}[t]
\centering
\includegraphics[width=0.99\textwidth]{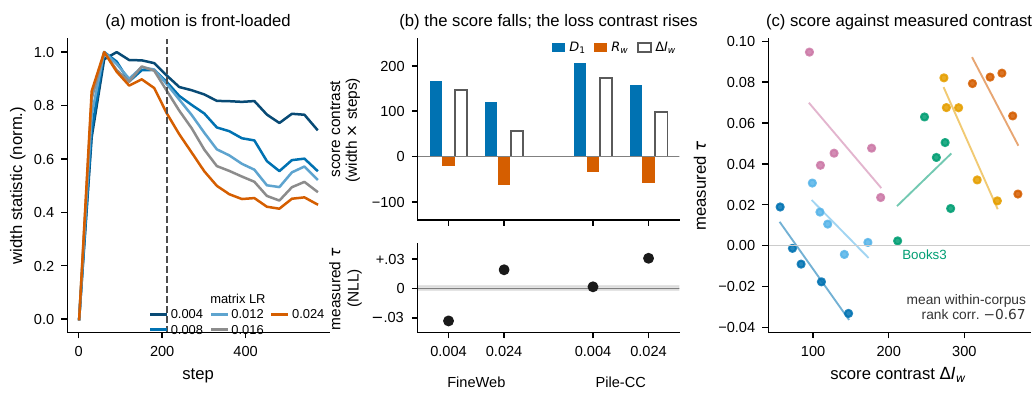}
\vspace{-0.2cm}
\caption{(a)~Normalised spectral-width profiles $w(t)$ of the 20-refresh
reference runs (FineWeb, five learning rates). The dashed line marks the
first step at which the two schedules differ. (b)~Top: decomposition of the
score contrast $\Delta I_w$ into $D_1$ and $R_w$ (Eq.~\ref{eq:gapsplit}), in
width $\times$ steps. Bottom: the
measured $\tauwin$ of the same cells, in NLL, with the tie band.
(c)~$\Delta I_w$ (Eq.~\ref{eq:score}) against measured $\tauwin$ over all
30 cells, with per-corpus fits.}
\label{fig:staleness}
\vspace{-0.2cm}
\end{figure*}

Error-triggered refresh rules rest on the idea that a refresh helps once
the cached curvature has drifted from the current training state. If that
idea explained placement, a measure of accumulated drift should predict
which schedule wins at each learning rate. We tested one measurable
version, accumulated change in the spectral width of the curvature
factors, as a registered prediction.

Let $w(t)$ be the spectral-width profile of a separately instrumented
reference run for the same (corpus, learning rate, seed) cell, which
refreshes at a set of probe steps. The statistic aggregates the log
ratio of upper and lower eigenvalue quantiles across matrix parameters
(App.~\ref{app:math}). Both schedules are evaluated on the same reference
profile. Using the last-refresh function of Eq.~\ref{eq:lastrefresh}, define
\begin{equation}
\begin{aligned}
I_w(S)&=\int_1^{H}\bigl|w(t)-w(r_S(t))\bigr|\,dt,\\
\Delta I_w&=I_w(S_L)-I_w(S_E).
\end{aligned}
\label{eq:score}
\end{equation}
A positive $\Delta I_w$ means that later placement accumulates more
spectral-width mismatch, so the score predicts an advantage for earlier
placement. The score is measured in spectral width times training steps.
It ranks schedules. The registered test compared the within-corpus rank order of $\Delta I_w$ with
that of the measured \tauwin across learning rates, against frozen targets
and a frozen withdrawal threshold (App.~\ref{app:mechanism}).

$\Delta I_w$ was positive in every cell, so it predicted earlier placement everywhere, while the measured
contrast changed sign on the crossing corpora. From the lowest to the
highest learning rate, $\Delta I_w$ fell in every corpus while the measured contrast rose in all
but one. The within-corpus rank correlation was negative on average and
positive only in Books3 (Figure~\ref{fig:staleness}c), whose contrast
collapses where its runs became unstable. The profiles themselves were stable across seeds
(App.~\ref{app:mechanism}). The registered prediction was
therefore withdrawn.

\paragraph{Source of the score.}
Most of the spectral-width motion occurred before the two schedules first differ
(Figure~\ref{fig:staleness}a). For the primary pair the score has the exact
decomposition
\begin{equation}
\Delta I_w=\underbrace{\int_{211}^{301}\bigl|w(t)-w(1)\bigr|\,dt}_{D_1}
+R_w ,
\label{eq:gapsplit}
\end{equation}
where $D_1$ is the extra exposure to the step-1 reference value under later
placement and $R_w$ collects the signed differences between the remaining
intervals (App.~\ref{app:math}). $D_1$ accounted for nearly all of the net contrast and often more, and $R_w$ was negative
in all but one cell.
Because $D_1$ compares the profile after step 211 with its value at step 1,
it mostly records curvature motion that happened before the schedules
diverge. Four other registered
operationalisations of staleness, including a single-interval drift
statistic, a data-freshness gain and a generalisation account, also failed
to predict the boundary (App.~\ref{app:mechanism}).

\paragraph{Spectral width and the update.}
Spectral width summarizes eigenvalue spread, while the preconditioned
update also depends on the eigenvectors and on the gradient expressed in
that basis. Two curvature factors related by a rotation have identical
spectral width, but their difference can be large in operator norm
(App.~\ref{app:math}). The score also measures drift on a common reference
trajectory, whereas training applies each cached factor along its own
schedule-specific trajectory. Front-loaded curvature motion has also been reported by
\citet{gurari2018subspace} and \citet{cohen2021edge}, and early training
is known to shape later training \citep{achille2019critical,frankle2020early}.

\section{Discussion and Conclusion}
\label{sec:conclusion}

We have shown that, in short language-model training with Mousse, a refresh interval leaves part of a refresh schedule free: at a fixed
budget, where the refreshes fall changes held-out loss, and which placement
wins depends on the learning rate, the model size and the training length.

Schedules should be compared across the learning rates at which they will
be used, because a single learning rate can report either ranking. Adding
one learning rate on each side of the tuned rate triples the cost of a
single-rate comparison and shows whether the ranking holds near the
operating point. An equal-budget uniform schedule makes a useful reference,
because two schedules at the same budget can both sit far from even
spacing, and a predictor of placement needs more than scalar drift of the
curvature spectrum.

Over the measured range the boundary fell faster with model size than the
loss-optimal learning rate, and the gap between them narrowed
(\S\ref{sec:transfer}). A learning rate tuned or transferred for a larger
model therefore sits closer to the boundary, where the two placements are
near a tie, so a schedule ranking obtained at one scale should be rechecked
after scaling.

Our results add a dimension to earlier evidence on refresh timing. SOAP
showed that longer refresh intervals degrade training
\citep{vyas2025soap}, and our experiments show that, at a fixed number of
refreshes, their placement matters as well. Purifying Shampoo found that
eigenbasis error matters far more in early iterations than in late ones
\citep{eschenhagen2025purifying}. Most spectral-width motion in our runs
also occurred early, before the two schedules diverge
(\S\ref{sec:staleness}), so the two observations agree that early
curvature carries most of the change.

Error-triggered refresh rules \citep{eschenhagen2025purifying,nam2026foam}
change when refreshes occur and are usually evaluated at one tuned learning rate, the setting to which these practices apply directly.

The natural next measurement records the cached eigenvectors at every
refresh and, at each probe step, computes their principal angles to the
fresh eigenbasis, weighted by how much of the gradient falls in each
direction. Such a score follows each schedule along its own trajectory
and sees rotation as well as spread, the two properties the spectral-width
score leaves out. Longer horizons and other Kronecker-preconditioned
optimizers are the other open tests.

\clearpage
\section*{Limitations}
\label{sec:limits}
\emph{One optimizer.} The reversal is established for Mousse. Restoring its
published internals preserved it, but attempts to extend the study to other
Kronecker-preconditioned optimizers were inconclusive, and the direction of
the effect may differ for other optimizers (App.~\ref{app:surface}).

\emph{Short training.} Runs use far fewer tokens per parameter than
compute-optimal training \citep{hoffmann2022chinchilla}, and the response
changes with horizon. Because model size and tokens per parameter vary
together, the cross-size boundary is a coarse trend over 64M to 929M.

\emph{Mechanism.} The spectral-width score and other staleness accounts
failed to predict the boundary, and measuring eigenvector rotation is left
to future work.

\section*{Ethics Statement}
This work measures optimizer dynamics on public pretraining corpora and
releases only code, registration protocols and adjudicated statistics. FineWeb is distributed under ODC-By 1.0. The Pile
subsets (GitHub, Pile-CC, PubMed~Central, arXiv, Books3) are used under
the Pile's terms as a compilation. Books3 contains copyrighted books, and
it appears here solely as one of six domain streams for a
training-dynamics measurement whose outputs are loss values. Its societal risks are those of language-model pretraining in general, and its practical recommendations concern
experimental reporting.

\bibliography{refs}

\begin{thebibliography}{36}
\providecommand{\natexlab}[1]{#1}

\bibitem[{Achille et~al.(2019)Achille, Rovere, and Soatto}]{achille2019critical}
Alessandro Achille, Matteo Rovere, and Stefano Soatto. 2019.
\newblock \href {https://openreview.net/forum?id=BkeStsCcKQ} {Critical learning periods in deep networks}.
\newblock In \emph{7th International Conference on Learning Representations, {ICLR} 2019, New Orleans, LA, USA, May 6-9, 2019}. OpenReview.net.

\bibitem[{Anil et~al.(2020)Anil, Gupta, Koren, Regan, and Singer}]{anil2020scalable}
Rohan Anil, Vineet Gupta, Tomer Koren, Kevin Regan, and Yoram Singer. 2020.
\newblock \href {https://arxiv.org/abs/2002.09018} {Scalable second order optimization for deep learning}.
\newblock \emph{CoRR}, abs/2002.09018.

\bibitem[{Bernstein and Newhouse(2024)}]{bernstein2024old}
Jeremy Bernstein and Laker Newhouse. 2024.
\newblock \href {https://doi.org/10.48550/ARXIV.2409.20325} {Old optimizer, new norm: An anthology}.
\newblock \emph{CoRR}, abs/2409.20325.

\bibitem[{Bjorck et~al.(2025)Bjorck, Benhaim, Chaudhary, Wei, and Song}]{bjorck2024tokenhorizon}
Johan Bjorck, Alon Benhaim, Vishrav Chaudhary, Furu Wei, and Xia Song. 2025.
\newblock \href {https://openreview.net/forum?id=WYL4eFLcxG} {Scaling optimal {LR} across token horizons}.
\newblock In \emph{The Thirteenth International Conference on Learning Representations, {ICLR} 2025, Singapore, April 24-28, 2025}. OpenReview.net.

\bibitem[{Chen et~al.(2023)Chen, Liang, Huang, Real, Wang, Liu, Pham, Dong, Luong, Hsieh, Lu, and Le}]{chen2023lion}
Xiangning Chen, Chen Liang, Da~Huang, Esteban Real, Kaiyuan Wang, Yao Liu, Hieu Pham, Xuanyi Dong, Thang Luong, Cho{-}Jui Hsieh, Yifeng Lu, and Quoc~V. Le. 2023.
\newblock \href {https://arxiv.org/abs/2302.06675} {Symbolic discovery of optimization algorithms}.
\newblock In \emph{Advances in Neural Information Processing Systems 36: Annual Conference on Neural Information Processing Systems 2023, NeurIPS 2023}.

\bibitem[{Choi et~al.(2019)Choi, Shallue, Nado, Lee, Maddison, and Dahl}]{choi2019empirical}
Dami Choi, Christopher~J. Shallue, Zachary Nado, Jaehoon Lee, Chris~J. Maddison, and George~E. Dahl. 2019.
\newblock \href {https://arxiv.org/abs/1910.05446} {On empirical comparisons of optimizers for deep learning}.
\newblock \emph{CoRR}, abs/1910.05446.

\bibitem[{Cohen et~al.(2021)Cohen, Kaur, Li, Kolter, and Talwalkar}]{cohen2021edge}
Jeremy Cohen, Simran Kaur, Yuanzhi Li, J.~Zico Kolter, and Ameet Talwalkar. 2021.
\newblock \href {https://openreview.net/forum?id=jh-rTtvkGeM} {Gradient descent on neural networks typically occurs at the edge of stability}.
\newblock In \emph{9th International Conference on Learning Representations, {ICLR} 2021, Virtual Event, Austria, May 3-7, 2021}. OpenReview.net.

\bibitem[{Dahl et~al.(2023)Dahl, Schneider, Nado, Agarwal, Sastry, Hennig, Medapati, Eschenhagen, Kasimbeg, Suo, Bae, Gilmer, Peirson, Khan, Anil, Rabbat, Krishnan, Snider, Amid, Chen, Maddison, Vasudev, Badura, Garg, and Mattson}]{dahl2023benchmarking}
George~E. Dahl, Frank Schneider, Zachary Nado, Naman Agarwal, Chandramouli~Shama Sastry, Philipp Hennig, Sourabh Medapati, Runa Eschenhagen, Priya Kasimbeg, Daniel Suo, Juhan Bae, Justin Gilmer, Abel~L. Peirson, Bilal Khan, Rohan Anil, Mike Rabbat, Shankar Krishnan, Daniel Snider, Ehsan Amid, and 6 others. 2023.
\newblock \href {https://doi.org/10.48550/ARXIV.2306.07179} {Benchmarking neural network training algorithms}.
\newblock \emph{CoRR}, abs/2306.07179.

\bibitem[{Eschenhagen et~al.(2025)Eschenhagen, Defazio, Lee, Turner, and Shi}]{eschenhagen2025purifying}
Runa Eschenhagen, Aaron Defazio, Tsung{-}Hsien Lee, Richard~E. Turner, and Hao{-}Jun~Michael Shi. 2025.
\newblock \href {http://papers.nips.cc/paper\_files/paper/2025/hash/f14f4eda29a74c02c803699a09529bb9-Abstract-Conference.html} {Purifying {Shampoo}: Investigating {Shampoo}'s heuristics by decomposing its preconditioner}.
\newblock In \emph{Advances in Neural Information Processing Systems 38: Annual Conference on Neural Information Processing Systems 2025, NeurIPS 2025, San Diego, CA, {USA}}.

\bibitem[{Frankle et~al.(2020)Frankle, Schwab, and Morcos}]{frankle2020early}
Jonathan Frankle, David~J. Schwab, and Ari~S. Morcos. 2020.
\newblock \href {https://openreview.net/forum?id=Hkl1iRNFwS} {The early phase of neural network training}.
\newblock In \emph{8th International Conference on Learning Representations, {ICLR} 2020, Addis Ababa, Ethiopia, April 26-30, 2020}. OpenReview.net.

\bibitem[{Gao et~al.(2021)Gao, Biderman, Black, Golding, Hoppe, Foster, Phang, He, Thite, Nabeshima, Presser, and Leahy}]{gao2020pile}
Leo Gao, Stella Biderman, Sid Black, Laurence Golding, Travis Hoppe, Charles Foster, Jason Phang, Horace He, Anish Thite, Noa Nabeshima, Shawn Presser, and Connor Leahy. 2021.
\newblock \href {https://arxiv.org/abs/2101.00027} {The {Pile}: An {800GB} dataset of diverse text for language modeling}.
\newblock \emph{CoRR}, abs/2101.00027.

\bibitem[{George et~al.(2018)George, Laurent, Bouthillier, Ballas, and Vincent}]{george2018ekfac}
Thomas George, C{\'{e}}sar Laurent, Xavier Bouthillier, Nicolas Ballas, and Pascal Vincent. 2018.
\newblock \href {https://proceedings.neurips.cc/paper/2018/hash/48000647b315f6f00f913caa757a70b3-Abstract.html} {Fast approximate natural gradient descent in a {Kronecker} factored eigenbasis}.
\newblock In \emph{Advances in Neural Information Processing Systems 31: Annual Conference on Neural Information Processing Systems 2018, NeurIPS 2018, December 3-8, 2018, Montr{\'{e}}al, Canada}.

\bibitem[{Gupta et~al.(2018)Gupta, Koren, and Singer}]{gupta2018shampoo}
Vineet Gupta, Tomer Koren, and Yoram Singer. 2018.
\newblock \href {http://proceedings.mlr.press/v80/gupta18a.html} {Shampoo: Preconditioned stochastic tensor optimization}.
\newblock In \emph{Proceedings of the 35th International Conference on Machine Learning, {ICML} 2018, Stockholmsm{\"{a}}ssan, Stockholm, Sweden, July 10-15, 2018}, volume~80 of \emph{Proceedings of Machine Learning Research}, pages 1842--1850. {PMLR}.

\bibitem[{Gur{-}Ari et~al.(2018)Gur{-}Ari, Roberts, and Dyer}]{gurari2018subspace}
Guy Gur{-}Ari, Daniel~A. Roberts, and Ethan Dyer. 2018.
\newblock \href {https://arxiv.org/abs/1812.04754} {Gradient descent happens in a tiny subspace}.
\newblock \emph{CoRR}, abs/1812.04754.

\bibitem[{H{\"{a}}gele et~al.(2024)H{\"{a}}gele, Bakouch, Kosson, Allal, von Werra, and Jaggi}]{hagele2024wsd}
Alexander H{\"{a}}gele, Elie Bakouch, Atli Kosson, Loubna~Ben Allal, Leandro von Werra, and Martin Jaggi. 2024.
\newblock \href {http://papers.nips.cc/paper\_files/paper/2024/hash/8b970e15a89bf5d12542810df8eae8fc-Abstract-Conference.html} {Scaling laws and compute-optimal training beyond fixed training durations}.
\newblock In \emph{Advances in Neural Information Processing Systems 37: Annual Conference on Neural Information Processing Systems 2024, NeurIPS 2024, Vancouver, BC, Canada, December 10 - 15, 2024}.

\bibitem[{Hoffmann et~al.(2022)Hoffmann, Borgeaud, Mensch, Buchatskaya, Cai, Rutherford, de~Las~Casas, Hendricks, Welbl, Clark, Hennigan, Noland, Millican, van~den Driessche, Damoc, Guy, Osindero, Simonyan, Elsen, Rae, Vinyals, and Sifre}]{hoffmann2022chinchilla}
Jordan Hoffmann, Sebastian Borgeaud, Arthur Mensch, Elena Buchatskaya, Trevor Cai, Eliza Rutherford, Diego de~Las~Casas, Lisa~Anne Hendricks, Johannes Welbl, Aidan Clark, Tom Hennigan, Eric Noland, Katie Millican, George van~den Driessche, Bogdan Damoc, Aurelia Guy, Simon Osindero, Karen Simonyan, Erich Elsen, and 3 others. 2022.
\newblock \href {https://arxiv.org/abs/2203.15556} {Training compute-optimal large language models}.
\newblock In \emph{Advances in Neural Information Processing Systems 35: Annual Conference on Neural Information Processing Systems 2022, NeurIPS 2022}.

\bibitem[{Hu et~al.(2024)Hu, Tu, Han, He, Cui, Long, Zheng, Fang, Huang, Zhao, Zhang, Thai, Zhang, Wang, Yao, Zhao, Zhou, Cai, Zhai, Ding, Jia, Zeng, Li, Liu, and Sun}]{hu2024minicpm}
Shengding Hu, Yuge Tu, Xu~Han, Chaoqun He, Ganqu Cui, Xiang Long, Zhi Zheng, Yewei Fang, Yuxiang Huang, Weilin Zhao, Xinrong Zhang, Zheng~Leng Thai, Kaihuo Zhang, Chongyi Wang, Yuan Yao, Chenyang Zhao, Jie Zhou, Jie Cai, Zhongwu Zhai, and 6 others. 2024.
\newblock \href {https://doi.org/10.48550/ARXIV.2404.06395} {{MiniCPM}: Unveiling the potential of small language models with scalable training strategies}.
\newblock \emph{CoRR}, abs/2404.06395.

\bibitem[{Jordan et~al.(2024)Jordan, Jin, Boza, You, Cesista, Newhouse, and Bernstein}]{jordan2024muon}
Keller Jordan, Yuchen Jin, Vlado Boza, Jiacheng You, Franz Cesista, Laker Newhouse, and Jeremy Bernstein. 2024.
\newblock {Muon}: An optimizer for hidden layers in neural networks.
\newblock \url{https://kellerjordan.github.io/posts/muon/}.

\bibitem[{Kasimbeg et~al.(2025)Kasimbeg, Schneider, Eschenhagen, Bae, Sastry, Saroufim, Feng, Wright, Yang, Nado, Medapati, Hennig, Rabbat, and Dahl}]{kasimbeg2025algoperf}
Priya Kasimbeg, Frank Schneider, Runa Eschenhagen, Juhan Bae, Chandramouli~Shama Sastry, Mark Saroufim, Boyuan Feng, Less Wright, Edward~Z. Yang, Zachary Nado, Sourabh Medapati, Philipp Hennig, Michael Rabbat, and George~E. Dahl. 2025.
\newblock Accelerating neural network training: An analysis of the {AlgoPerf} competition.
\newblock In \emph{The Thirteenth International Conference on Learning Representations, {ICLR} 2025, Singapore, April 24-28, 2025}. OpenReview.net.

\bibitem[{Liu et~al.(2025)Liu, Su, Yao, Jiang, Lai, Du, Qin, Xu, Lu, Yan, Chen, Zheng, Liu, Liu, Yin, He, Zhu, Wang, Wang, Dong, Zhang, Kang, Zhang, Xu, Zhang, Wu, Zhou, and Yang}]{liu2025muon}
Jingyuan Liu, Jianlin Su, Xingcheng Yao, Zhejun Jiang, Guokun Lai, Yulun Du, Yidao Qin, Weixin Xu, Enzhe Lu, Junjie Yan, Yanru Chen, Huabin Zheng, Yibo Liu, Shaowei Liu, Bohong Yin, Weiran He, Han Zhu, Yuzhi Wang, Jianzhou Wang, and 9 others. 2025.
\newblock \href {https://doi.org/10.48550/ARXIV.2502.16982} {Muon is scalable for {LLM} training}.
\newblock \emph{CoRR}, abs/2502.16982.

\bibitem[{Loshchilov and Hutter(2019)}]{loshchilov2019adamw}
Ilya Loshchilov and Frank Hutter. 2019.
\newblock \href {https://openreview.net/forum?id=Bkg6RiCqY7} {Decoupled weight decay regularization}.
\newblock In \emph{7th International Conference on Learning Representations, {ICLR} 2019, New Orleans, LA, USA, May 6-9, 2019}. OpenReview.net.

\bibitem[{Martens and Grosse(2015)}]{martens2015kfac}
James Martens and Roger Grosse. 2015.
\newblock \href {http://proceedings.mlr.press/v37/martens15.html} {Optimizing neural networks with {Kronecker}-factored approximate curvature}.
\newblock In \emph{Proceedings of the 32nd International Conference on Machine Learning, {ICML} 2015, Lille, France, 6-11 July 2015}, volume~37 of \emph{{JMLR} Workshop and Conference Proceedings}, pages 2408--2417. JMLR.org.

\bibitem[{Morwani et~al.(2025)Morwani, Shapira, Vyas, Malach, Kakade, and Janson}]{morwani2025newperspective}
Depen Morwani, Itai Shapira, Nikhil Vyas, Eran Malach, Sham~M. Kakade, and Lucas Janson. 2025.
\newblock \href {https://arxiv.org/abs/2406.17748} {A new perspective on {Shampoo}'s preconditioner}.
\newblock In \emph{The Thirteenth International Conference on Learning Representations, {ICLR} 2025, Singapore, April 24-28, 2025}. OpenReview.net.

\bibitem[{Nam and Ahn(2026)}]{nam2026foam}
Kyunghun Nam and Sumyeong Ahn. 2026.
\newblock \href {https://arxiv.org/abs/2606.02365} {{FOAM}: Frequency and operator error-based adaptive damping method for reducing staleness-oriented error for {Shampoo}}.
\newblock \emph{CoRR}, abs/2606.02365.

\bibitem[{Paszke et~al.(2019)Paszke, Gross, Massa, Lerer, Bradbury, Chanan, Killeen, Lin, Gimelshein, Antiga, Desmaison, K{\"{o}}pf, Yang, DeVito, Raison, Tejani, Chilamkurthy, Steiner, Fang, Bai, and Chintala}]{paszke2019pytorch}
Adam Paszke, Sam Gross, Francisco Massa, Adam Lerer, James Bradbury, Gregory Chanan, Trevor Killeen, Zeming Lin, Natalia Gimelshein, Luca Antiga, Alban Desmaison, Andreas K{\"{o}}pf, Edward Yang, Zachary DeVito, Martin Raison, Alykhan Tejani, Sasank Chilamkurthy, Benoit Steiner, Lu~Fang, and 2 others. 2019.
\newblock \href {https://arxiv.org/abs/1912.01703} {{PyTorch}: An imperative style, high-performance deep learning library}.
\newblock In \emph{Advances in Neural Information Processing Systems 32: Annual Conference on Neural Information Processing Systems 2019, NeurIPS 2019}.

\bibitem[{Penedo et~al.(2024)Penedo, Kydl{\'\i}{\v{c}}ek, Allal, Lozhkov, Mitchell, Raffel, von Werra, and Wolf}]{penedo2024fineweb}
Guilherme Penedo, Hynek Kydl{\'\i}{\v{c}}ek, Loubna~Ben Allal, Anton Lozhkov, Margaret Mitchell, Colin Raffel, Leandro von Werra, and Thomas Wolf. 2024.
\newblock \href {http://papers.nips.cc/paper\_files/paper/2024/hash/370df50ccfdf8bde18f8f9c2d9151bda-Abstract-Datasets\_and\_Benchmarks\_Track.html} {The {FineWeb} datasets: Decanting the web for the finest text data at scale}.
\newblock In \emph{Advances in Neural Information Processing Systems 37: Annual Conference on Neural Information Processing Systems 2024, NeurIPS 2024, Vancouver, BC, Canada, December 10 - 15, 2024}.

\bibitem[{Radford et~al.(2019)Radford, Wu, Child, Luan, Amodei, and Sutskever}]{radford2019gpt2}
Alec Radford, Jeffrey Wu, Rewon Child, David Luan, Dario Amodei, and Ilya Sutskever. 2019.
\newblock Language models are unsupervised multitask learners.
\newblock Technical report, OpenAI.

\bibitem[{Schmidt et~al.(2021)Schmidt, Schneider, and Hennig}]{schmidt2021crowded}
Robin~M. Schmidt, Frank Schneider, and Philipp Hennig. 2021.
\newblock \href {http://proceedings.mlr.press/v139/schmidt21a.html} {Descending through a crowded valley - benchmarking deep learning optimizers}.
\newblock In \emph{Proceedings of the 38th International Conference on Machine Learning, {ICML} 2021, 18-24 July 2021, Virtual Event}, volume 139 of \emph{Proceedings of Machine Learning Research}, pages 9367--9376. {PMLR}.

\bibitem[{Semenov et~al.(2025)Semenov, Pagliardini, and Jaggi}]{semenov2025benchmarking}
Andrei Semenov, Matteo Pagliardini, and Martin Jaggi. 2025.
\newblock \href {https://doi.org/10.48550/ARXIV.2509.01440} {Benchmarking optimizers for large language model pretraining}.
\newblock \emph{CoRR}, abs/2509.01440.

\bibitem[{Shi et~al.(2023)Shi, Lee, Iwasaki, Gallego{-}Posada, Li, Rangadurai, Mudigere, and Rabbat}]{shi2023shampoo}
Hao{-}Jun~Michael Shi, Tsung{-}Hsien Lee, Shintaro Iwasaki, Jose Gallego{-}Posada, Zhijing Li, Kaushik Rangadurai, Dheevatsa Mudigere, and Michael Rabbat. 2023.
\newblock \href {https://doi.org/10.48550/ARXIV.2309.06497} {A distributed data-parallel {PyTorch} implementation of the distributed {Shampoo} optimizer for training neural networks at-scale}.
\newblock \emph{CoRR}, abs/2309.06497.

\bibitem[{Sivaprasad et~al.(2020)Sivaprasad, Mai, Vogels, Jaggi, and Fleuret}]{sivaprasad2020optimizer}
Prabhu~Teja Sivaprasad, Florian Mai, Thijs Vogels, Martin Jaggi, and Fran{\c{c}}ois Fleuret. 2020.
\newblock \href {http://proceedings.mlr.press/v119/sivaprasad20a.html} {Optimizer benchmarking needs to account for hyperparameter tuning}.
\newblock In \emph{Proceedings of the 37th International Conference on Machine Learning, {ICML} 2020, 13-18 July 2020, Virtual Event}, volume 119 of \emph{Proceedings of Machine Learning Research}, pages 9036--9045. {PMLR}.

\bibitem[{Vyas et~al.(2025)Vyas, Morwani, Zhao, Shapira, Brandfonbrener, Janson, and Kakade}]{vyas2025soap}
Nikhil Vyas, Depen Morwani, Rosie Zhao, Itai Shapira, David Brandfonbrener, Lucas Janson, and Sham~M. Kakade. 2025.
\newblock \href {https://openreview.net/forum?id=IDxZhXrpNf} {{SOAP}: Improving and stabilizing {Shampoo} using {Adam} for language modeling}.
\newblock In \emph{The Thirteenth International Conference on Learning Representations, {ICLR} 2025, Singapore, April 24-28, 2025}. OpenReview.net.

\bibitem[{Wen et~al.(2026)Wen, Hall, Ma, and Liang}]{wen2025fantastic}
Kaiyue Wen, David Hall, Tengyu Ma, and Percy Liang. 2026.
\newblock \href {https://arxiv.org/abs/2509.02046} {Fantastic pretraining optimizers and where to find them}.
\newblock In \emph{The Fourteenth International Conference on Learning Representations, {ICLR} 2026}.

\bibitem[{Yang et~al.(2021)Yang, Hu, Babuschkin, Sidor, Liu, Farhi, Ryder, Pachocki, Chen, and Gao}]{yang2022mup}
Greg Yang, Edward~J. Hu, Igor Babuschkin, Szymon Sidor, Xiaodong Liu, David Farhi, Nick Ryder, Jakub Pachocki, Weizhu Chen, and Jianfeng Gao. 2021.
\newblock \href {https://arxiv.org/abs/2203.03466} {Tuning large neural networks via zero-shot hyperparameter transfer}.
\newblock In \emph{Advances in Neural Information Processing Systems 34: Annual Conference on Neural Information Processing Systems 2021, NeurIPS 2021}.

\bibitem[{Zhang et~al.(2026)Zhang, Xing, Huang, Lv, Zhou, Qiu, Guo, and Chen}]{zhang2026mousse}
Yechen Zhang, Shuhao Xing, Junhao Huang, Kai Lv, Yunhua Zhou, Xipeng Qiu, Qipeng Guo, and Kai Chen. 2026.
\newblock \href {https://doi.org/10.48550/ARXIV.2603.09697} {Mousse: Rectifying the geometry of {Muon} with curvature-aware preconditioning}.
\newblock \emph{CoRR}, abs/2603.09697.

\bibitem[{Zhao et~al.(2025)Zhao, Morwani, Brandfonbrener, Vyas, and Kakade}]{zhao2025deconstructing}
Rosie Zhao, Depen Morwani, David Brandfonbrener, Nikhil Vyas, and Sham~M. Kakade. 2025.
\newblock \href {https://arxiv.org/abs/2407.07972} {Deconstructing what makes a good optimizer for autoregressive language models}.
\newblock In \emph{The Thirteenth International Conference on Learning Representations, {ICLR} 2025, Singapore, April 24-28, 2025}. OpenReview.net.

\end{thebibliography}

\appendix
\renewcommand{\theequation}{A\arabic{equation}}\renewcommand{\theHequation}{A\arabic{equation}}\setcounter{equation}{0}
\section{Derivations}
\label{app:math}

\paragraph{Cumulative cache age.}
For updates $t=1,\ldots,H$ and a schedule $S=(s_1,\ldots,s_K)$ with
$s_1=1$, set $s_{K+1}=H+1$ and $g_j=s_{j+1}-s_j$. On the updates
$s_j,\ldots,s_{j+1}-1$ the cached factorization has ages $0,\ldots,g_j-1$,
so summing the arithmetic progression on each interval gives
\begin{equation}
\sum_{t=1}^{H} a_S(t)=\tfrac12\sum_{j=1}^{K} g_j\,(g_j-1).
\label{eq:cumage}
\end{equation}
The sum is convex in the gaps, so with the endpoints fixed it is smallest
when the free gaps are as equal as possible. For the primary schedules at
$H=591$ the gaps are $g_E=(210,80,270,31)$ and $g_L=(300,80,180,31)$,
giving cumulative ages of $61{,}885$ and $64{,}585$. The two schedules share
the refresh count and the terminal age, but their cumulative ages differ by
$2{,}700$ step-age units. Cumulative age is one of several quantities that change with placement. A refresh at a common
step can also produce different caches, because earlier placement
differences change the parameters, gradients, momentum and curvature
statistics from which that refresh is computed.

\paragraph{Interpolated preference boundary.}
For an ordered learning-rate grid $\eta_1<\cdots<\eta_m$, we took adjacent
points whose seed-mean contrasts have opposite signs, interpolated $\bar\tau$
linearly in $x=\log\eta$ inside the selected bracket $(\eta_j,\eta_{j+1})$,
and set the interpolant to zero:
\begin{equation}
\log\lrx=\log\eta_j-\frac{\bar\tau(\eta_j)}
{\bar\tau(\eta_{j+1})-\bar\tau(\eta_j)}\,\log\frac{\eta_{j+1}}{\eta_j}.
\label{eq:crossing}
\end{equation}
When a grid contains several sign changes we used the last bracket, and we kept every estimate inside the tested grid. The estimate is a point inside
an observed sign-change interval. It needs only one observed sign change. The contrast may be non-monotone,
have several zeros, or cross elsewhere under a new seed.

\paragraph{Gap-term decomposition of the spectral-width score.}
For a reference profile $w$ on $[1,H]$, write $d_w(t,s)=|w(t)-w(s)|$ and
treat $r_S$ as a right-continuous step function on $[1,H]$. Expanding
Eq.~\ref{eq:score} over the primary schedules gives
\begin{align*}
I_w(S_E)={}&\int_1^{211}\! d_w(t,1)\,dt+\int_{211}^{291}\! d_w(t,211)\,dt\\
&+\int_{291}^{561}\! d_w(t,291)\,dt\\
&+\int_{561}^{H}\! d_w(t,561)\,dt,\\[2pt]
I_w(S_L)={}&\int_1^{301}\! d_w(t,1)\,dt+\int_{301}^{381}\! d_w(t,301)\,dt\\
&+\int_{381}^{561}\! d_w(t,381)\,dt\\
&+\int_{561}^{H}\! d_w(t,561)\,dt,
\end{align*}
and subtracting gives the exact identity
\begin{equation}
\Delta I_w=D_1+D_2+D_3,
\label{eq:gapterms}
\end{equation}
with
\begin{align*}
D_1&=\int_{211}^{301} d_w(t,1)\,dt,\\
D_2&=\int_{301}^{381} d_w(t,301)\,dt-\int_{211}^{291} d_w(t,211)\,dt,\\
D_3&=\int_{381}^{561} d_w(t,381)\,dt-\int_{291}^{561} d_w(t,291)\,dt,
\end{align*}
so that $R_w=D_2+D_3$ in Eq.~\ref{eq:gapsplit}. The shared first and last
intervals cancel because both scores use the same profile. The identity concerns the diagnostic only. Loss effects along the two
training trajectories are a separate quantity. $D_1$ is non-negative while
$D_2$ and $D_3$ are signed, so $\Delta I_w$ can take either sign, and a
share $D_1/\Delta I_w>1$ means that $R_w$ is negative.

For each matrix parameter the width is
$w_{\mathrm{mat}}(t)=\log_{10}\!\big(q_{95}(\Lambda_L(t))/q_{05}(\Lambda_L(t))\big)
+\log_{10}\!\big(q_{95}(\Lambda_R(t))/q_{05}(\Lambda_R(t))\big)$
on the applied eigenvalues. The run-level profile is the median over matrix
parameters, interpolated linearly between probes and held at its last value
after the final probe. The integrals were evaluated by the trapezoid rule on
the unit step grid for each seed and averaged over the two seeds of a cell.

\paragraph{Spectral width and eigenvector orientation.}
An eigenspectrum leaves the orientation of its eigenvectors free.
For $\kappa>1$ let $A=\mathrm{diag}(1,\kappa)$ and $B=\mathrm{diag}(\kappa,1)$.
The two matrices have the same eigenvalues, so every width statistic built
from eigenvalue quantiles takes the same value on both, but
$\|A-B\|_2=\kappa-1$ and $B=RAR^{\!\top}$ for the rotation
$R=\bigl(\begin{smallmatrix}0&-1\\1&0\end{smallmatrix}\bigr)$. Zero change
in spectral width can therefore coexist with a large change in the
curvature matrix. The example shows a limit of the statistic. Whether such rotations occur in
our runs, and whether they relate to the reversal, is open
(App.~\ref{app:mechanism}). Similarly, a falling unweighted score can
coexist with a rising weighted one:
components that move from $(10,1)$ to $(7,2)$ have a falling unweighted sum
but a rising sum under weights $(1,4)$.

\section{Implementation and Training Details}
\label{app:impl}

\paragraph{Optimizer update.}
Matrix-shaped parameters (attention and MLP weights, $48$ to $128$ matrices
per model) were updated with the Mousse update of \citet{zhang2026mousse},
and embeddings, norms and other non-matrix parameters with AdamW. For one
matrix parameter $W\in\mathbb{R}^{m\times n}$ with gradient $G_t$, every
step maintains a heavy-ball momentum buffer and two exponential moving
averages of factored curvature:
\begin{equation}
\begin{aligned}
M_t &= \mu M_{t-1} + G_t,\\
L_t &= \beta L_{t-1} + (1{-}\beta)\, G_t G_t^{\!\top},\\
R_t &= \beta R_{t-1} + (1{-}\beta)\, G_t^{\!\top} G_t,
\end{aligned}
\label{eq:ema}
\end{equation}
with $\mu{=}0.95$ and $\beta{=}0.95$, called EMAs below. A refresh at a step $s\in S$ applies
bias correction and trace normalization
($\hat L \leftarrow m\hat L/\operatorname{tr}\hat L$), computes
$\operatorname{eigh}(\hat L + \varepsilon I) = Q_L \Lambda_L Q_L^{\!\top}$,
shifts the eigenvalues as
$\lambda \leftarrow \lambda + \max\{0,-\lambda_{\min}\} + \varepsilon$, and
caches $(\Lambda_L, Q_L)$, and likewise $(\Lambda_R, Q_R)$. Between refreshes
the moving averages keep evolving, but every update applies the cached
eigensystem, whose age is $a_S(t)$ of Eq.~\ref{eq:lastrefresh}. Each step
preconditions the momentum with the cached eigensystem and then
orthogonalizes it:
\begin{equation}
\begin{aligned}
\tilde M &= \Lambda_L^{-\alpha} Q_L^{\!\top} M_t\, Q_R
\Lambda_R^{-\alpha},\\
O &= \mathrm{NS5}(\tilde M),\\
\hat O &= Q_L \Lambda_L^{-\alpha} O\, \Lambda_R^{-\alpha}
Q_R^{\!\top},
\end{aligned}
\label{eq:update}
\end{equation}
where $\mathrm{NS5}$ is the five-iteration Newton--Schulz map and
$\alpha=0.125$ is applied before and after it. The update is rescaled to the Frobenius norm of $O$
($\hat O \leftarrow \hat O\,\|O\|_F/\|\hat O\|_F$) and applied with
decoupled weight decay $\omega$ and a spectral-norm learning-rate
adjustment:
\begin{equation}
W \leftarrow (1 - \eta\,\omega)\,W \;-\;
\eta\sqrt{m/n}\;\hat O.
\label{eq:apply}
\end{equation}
Algorithm~\ref{alg:mousse} summarizes one step.

\begin{algorithm}[t]
\small
\caption{One optimizer step (matrix parameters)}
\label{alg:mousse}
\begin{algorithmic}[1]
\State $M \gets \mu M + G$;\quad
$L \gets \beta L + (1{-}\beta) G G^{\top}$;\quad
$R \gets \beta R + (1{-}\beta) G^{\top} G$
\If{$t \in S$ \Comment{refresh schedule}}
\State $(\Lambda_L, Q_L) \gets
\mathrm{eigh}(\mathrm{tracenorm}(\mathrm{biascorr}(L)) + \varepsilon I)$;
shift $\lambda \gets \lambda + \max\{0,-\lambda_{\min}\} + \varepsilon$; same for $R$
\EndIf
\State $\tilde M \gets \Lambda_L^{-\alpha} Q_L^{\top} M Q_R
\Lambda_R^{-\alpha}$
\Comment{cached $(\Lambda, Q)$ from the \emph{last refresh}}
\State $O \gets \mathrm{NS5}(\tilde M)$;\quad
$\hat O \gets Q_L \Lambda_L^{-\alpha} O \Lambda_R^{-\alpha}
Q_R^{\top}$;\quad
$\hat O \gets \hat O\, \|O\|_F / \|\hat O\|_F$
\State $W \gets (1 - \eta\omega) W - \eta\sqrt{m/n}\,\hat O$
\end{algorithmic}
\end{algorithm}

\begin{table}[t]
\centering\footnotesize
\renewcommand{\arraystretch}{1.12}
\caption{Hyperparameters used in all experiments unless stated otherwise.
Matrix learning-rate grids are given with each experiment.}
\label{tab:hyper}
\begin{tabular}{@{}p{0.34\columnwidth}p{0.58\columnwidth}@{}}
\toprule
\multicolumn{2}{@{}l}{\emph{Optimizer (matrix parameters)}} \\
momentum $\mu$ & $0.95$ (heavy ball, Nesterov off) \\
curvature EMA $\beta$ & $0.95$, bias-corrected at refresh \\
curvature exponent $\alpha$ & 0.125, before and after Newton--Schulz \\
eigenvalue cleanup $\varepsilon$ & $10^{-10}$ (nonnegative correction
after trace normalization) \\
orthogonalization & Newton--Schulz, 5 iterations \\
norm grafting & update rescaled to $\|O\|_F$ \\
LR adjustment & spectral-norm, $\eta\sqrt{m/n}$ \\
\midrule
\multicolumn{2}{@{}l}{\emph{Non-matrix parameters}} \\
optimizer & AdamW, lr $3{\times}10^{-4}$, $\beta=(0.9, 0.95)$ \\
\midrule
\multicolumn{2}{@{}l}{\emph{Training}} \\
LR schedule & WSD: 10\% warmup, constant, 10\% warmdown \\
weight decay & $0.01$, decoupled ($0.0$ in the 64M boundary experiment) \\
sequence / tokens per step & $1{,}024$ / $65{,}536$ \\
batch geometry & $8{\times}8$ (64M/124M), $4{\times}16$ (350M),
$1{\times}64$ (695M/929M) micro$\times$accum \\
precision & bf16 autocast, fp32 curvature statistics \\
\bottomrule
\end{tabular}

\end{table}

\paragraph{Differences from published Mousse.}
We implemented the published update: whitening, Newton--Schulz
orthogonalization and unwhitening, trace normalization, $\beta=0.95$,
$\alpha=0.125$ and Frobenius-norm grafting. The intended change is the
refresh schedule, which replaces the published uniform interval of ten
steps. Three further internals differ: we shift the spectrum as
$\lambda\leftarrow\lambda+\max\{0,-\lambda_{\min}\}+10^{-10}$ instead of
adding $10^{-5}I$ damping, we apply a $\sqrt{m/n}$ matrix learning-rate
adjustment, and we optimize non-matrix parameters with AdamW instead of
Lion. App.~\ref{app:surface} restores all three. Arbitrary placement is
implemented by setting the refresh interval to a value larger than the
horizon and forcing a refresh exactly at the schedule steps. A per-step
check verified the age of the applied cache in every run.

\paragraph{Models, evaluation and compute.}
All models are GPT-style decoders with vocabulary 50{,}304 and context
1{,}024: 64M has 12 layers of width 512 and 8 heads, 124M has 12 layers of
width 768 and 12 heads, 350M has 24 layers of width 1{,}024 and 16 heads,
695M has 32 layers of width 1{,}280 and 20 heads, and 929M has 30 layers of
width 1{,}536 and 24 heads. Every evaluation uses the same 2{,}097{,}152-token held-out stream from a
shard disjoint from training. A run counts as unstable when its endpoint
NLL exceeds the best endpoint of its condition by more than $0.15$, the
stability margin.
Each run used one A100-SXM4-40GB from a pool of eight identical devices.
Median training time per run at 591 steps, excluding evaluation, was 6.2
(64M), 7.5 (124M), 18.5 (350M), 54.6 (695M) and 55.4 minutes (929M). The
experiments comprise $1{,}973$ completed runs and about $920$ A100-hours of
training. Software: PyTorch 2.7.1 \citep{paszke2019pytorch}, CUDA 12.6, Python 3.11.

\paragraph{Pre-specification and reproducibility.}
\label{app:registration}
For each experiment, the protocol, learning-rate grid, seeds, decision rule
and analysis script were fixed before the experiment ran. Seeds were fresh
for every experiment, and results from different experiments were kept separate. Experiments were designed in sequence, each after the results of earlier
ones were known. Training is
bit-deterministic given the seed: six duplicate configurations in the
learning-rate-ratio experiment, run separately and in two cases on different
devices, reproduced endpoint NLL exactly. The code, protocols and analysis
scripts are provided in anonymized form as supplementary material.

\section{Additional Results for Section~\ref{sec:reversal}}
\label{app:surface}

\paragraph{Tie tolerance and seed variation.}
The tie tolerance $\delta=0.002$ was fixed before the first surface
experiment. On the surface the median disagreement between the two seeds of
a cell was $.0045$ (90th percentile $.021$). Taking half that median, $.00225$, as a unit of seed variation, the
tolerance is about one unit, so a cell whose true contrast is zero would
often fall outside it with two seeds. The FineWeb contrasts on either
side of the reversal, $-.009$ and $+.019$, are about four and eight units
from zero. Two near-zero cells are used only descriptively: the Pile-CC dip
at $.008$ ($-.0043$, about two units) and the 124M tie at $.024$, which an
independent 124M cohort measured at $-.0067$.

\begin{figure}[h]
\centering
\includegraphics[width=0.9\columnwidth]{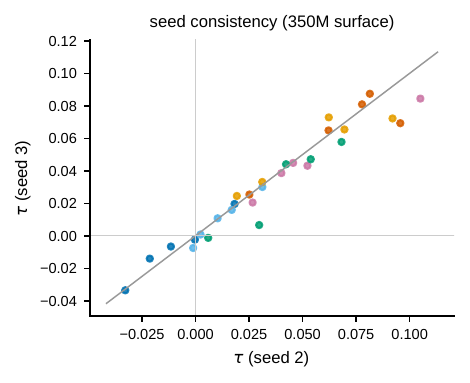}
\caption{Per-cell $\tauwin$ from the two seeds of the 350M surface, one
point per (corpus, LR) cell. The line is $y=x$.}
\label{fig:seedconsistency}
\end{figure}

\begin{table}[h]
\centering\footnotesize
\setlength{\tabcolsep}{4.5pt}\renewcommand{\arraystretch}{1.12}
\caption{Seed-mean $\tauwin$ at 350M and 591 steps (positive $=$ early
better).}
\label{tab:surface}
\begin{tabular}{@{}lrrrrr@{}}
\toprule
corpus & .004 & .008 & .012 & .016 & .024 \\
\midrule
FineWeb & $-.0332$ & $-.0177$ & $-.0091$ & $-.0013$ & $+.0189$ \\
Pile-CC & $+.0016$ & $-.0043$ & $+.0105$ & $+.0164$ & $+.0306$ \\
GitHub & $+.0253$ & $+.0635$ & $+.0844$ & $+.0825$ & $+.0793$ \\
PubMed & $+.0219$ & $+.0321$ & $+.0675$ & $+.0676$ & $+.0821$ \\
arXiv & $+.0236$ & $+.0477$ & $+.0452$ & $+.0393$ & $+.0947$ \\
Books3 & $+.0182$ & $+.0505$ & $+.0432$ & $+.0630$ & $+.0023$ \\
\bottomrule
\end{tabular}

\end{table}

\paragraph{Full surface.}
Table~\ref{tab:surface} gives every cell of Figure~\ref{fig:surface}, and
all 30 cells were seed-consistent (Figure~\ref{fig:seedconsistency}). At $.024$ the Books3 runs exceeded the stability margin. A dedicated
Books3 experiment measured $+.0632$ at $.016$ and $+.0038$ at $.024$.

\paragraph{Tuned operating points and the uniform reference.}
\label{app:dominance}
The tuned learning rate in Table~\ref{tab:tuned} is the grid value that
minimizes endpoint NLL for the better of the two mirrored schedules. At
FineWeb's tuned rate the effect is $.009$, about $15\%$ of the loss
reduction from doubling the model from 350M to 695M. At FineWeb's largest
surface effect, at $.004$, it is $57\%$ of that reduction, but this rate is
about three times below the optimum of the uniform reference. At the tuned
rates, the uniform schedule $(1,188,374,561)$ beat the better mirrored
schedule on five of the six corpora, by $.018$ to $.051$ NLL (three seeds).

\paragraph{Larger refresh budgets.}
We repeated the mirrored contrast at budgets 8 and 16 (FineWeb, 350M, three
seeds, 60 runs), keeping the first, last and window-end refreshes and
filling the windows with evenly spaced refreshes, a rule that reproduces the
primary pair at budget 4. The medians over the five surface learning rates
were $-.0395,-.0237,-.0084,-.0019,+.0188$ at budget 8 (crossing $.0166$)
and $-.0405,-.0232,-.0088,-.0036,+.0184$ at budget 16 (crossing $.0171$),
against $.0166$ at budget 4, and the contrast kept its size. This
construction keeps the first gap at 210 or 300 steps and the largest gap at
270 or 300 steps at every budget.

\begin{figure}[t]
\centering
\includegraphics[width=0.98\columnwidth]{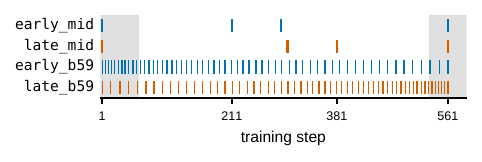}
\caption{The schedules as tick marks (grey bands are WSD warmup and
warmdown). Top: the primary mirrored pair. Bottom: the budget-59 pair,
whose interval lengths run in opposite directions while every gap stays
between 5 and 15 steps. The labels early\_mid and late\_mid denote $S_E$
and $S_L$, and early\_b59 and late\_b59 the budget-59 pair.}
\label{fig:schedules}
\end{figure}

\paragraph{Published refresh density.}
The budget-59 pair, the published density at this horizon, keeps every gap
short. Its 58 intervals form an arithmetic
progression that sums to 560 with largest-to-smallest ratio 3. The early
schedule takes the intervals in ascending order and the late schedule in
descending order. Both refresh first at step 1 and last at step 561, so
they share the terminal cache age of 30. Every interval lies between 5 and
15 steps (mean 9.66). We ran five learning rates with three fresh seeds,
and each matched pair ran back-to-back on one device.
Table~\ref{tab:dense} gives the results and Figure~\ref{fig:schedules}
draws the schedules.

\begin{table}[t]
\centering\footnotesize
\setlength{\tabcolsep}{5pt}
\caption{$\tauwin$ at budget 59 (FineWeb, 350M, three seeds). A cell is
consistent when all seeds receive the same label (early better, late
better, or tie).}
\label{tab:dense}
\begin{tabular}{@{}lrrrrc@{}}
\toprule
LR & seed 1 & seed 2 & seed 3 & median & consistent \\
\midrule
$.008$ & $-.0080$ & $-.0067$ & $-.0077$ & $-.0077$ & yes \\
$.012$ & $-.0048$ & $-.0054$ & $-.0042$ & $-.0048$ & yes \\
$.016$ & $-.0065$ & $-.0153$ & $-.0037$ & $-.0065$ & yes \\
$.020$ & $-.0007$ & $-.0079$ & $+.0018$ & $-.0007$ & no \\
$.024$ & $+.0069$ & $+.0144$ & $-.0028$ & $+.0069$ & no \\
\bottomrule
\end{tabular}

\end{table}

\paragraph{Placement landscape.}
The landscape of Figure~\ref{fig:landscape} uses $(1, c{-}40, c{+}40, 561)$
with $c\in\{251,281,296,311,341\}$. The budget, the first and last refresh,
the terminal cache age and the 80-step separation are fixed, so $c$ is the
only free coordinate. Reflecting the middle pair about step 296 maps the set
onto itself: $251$ and $341$ are the primary pair, $281$ and $311$ are an
interior pair, and $296$ is fixed. We ran three learning rates with three
fresh seeds (45 runs), with the five schedules for each (learning rate,
seed) back-to-back on one device. Adjacent centres differed by up to
$.0103$ at $.024$, while the best sampled centre and its neighbours
differed by $.0007$ to $.0025$.

\paragraph{Auxiliary learning-rate ratio.}
Sweeping the matrix learning rate also sweeps its ratio to the AdamW rate
of the non-matrix parameters, which the surface holds at $3\times10^{-4}$.
We separated the two on the primary pair with three fresh seeds. The fixed
condition kept AdamW at $3\times10^{-4}$, and the proportional condition set
it to $r\eta$ with $r=3\times10^{-4}/.016$, so that the ratio is constant
across learning rates. The fixed condition served as a positive control and
reproduced the bracket. Table~\ref{tab:ratio} gives both conditions.

\begin{table}[t]
\centering\footnotesize
\setlength{\tabcolsep}{3.1pt}
\caption{$\tauwin$ with the AdamW rate fixed at $3\times10^{-4}$ versus
scaled with the matrix rate (FineWeb, 350M, primary pair, three seeds). The
two conditions coincide at $.016$ by construction.}
\label{tab:ratio}
\begin{tabular}{@{}llrrrrr@{}}
\toprule
cond. & LR & AdamW & seed 1 & seed 2 & seed 3 & median \\
\midrule
fixed & $.012$ & $3.0$e$-4$ & $-.0060$ & $-.0049$ & $-.0046$ & $-.0049$ \\
fixed & $.016$ & $3.0$e$-4$ & $+.0005$ & $-.0007$ & $+.0033$ & $+.0005$ \\
fixed & $.024$ & $3.0$e$-4$ & $+.0157$ & $+.0141$ & $+.0163$ & $+.0157$ \\
\midrule
prop. & $.012$ & $2.25$e$-4$ & $-.0056$ & $-.0045$ & $-.0044$ & $-.0045$ \\
prop. & $.016$ & $3.0$e$-4$ & $+.0005$ & $-.0007$ & $+.0033$ & $+.0005$ \\
prop. & $.024$ & $4.5$e$-4$ & $+.0204$ & $+.0197$ & $+.0181$ & $+.0197$ \\
\bottomrule
\end{tabular}

\end{table}

\paragraph{Warmup-end refresh.}
Both schedules additionally refreshed at step 59, the end of warmup (FineWeb
and Pile-CC, five learning rates, three fresh seeds, 60 runs). The FineWeb
crossing moved from $.0166$ to $.0164$. On Pile-CC the bracket was present,
but this cohort differed from the surface by more than the tolerance at
$.004$ ($-.0213$ against $+.0016$) and at $.012$ ($-.0102$ against
$+.0105$), and its crossing ($.0152$) sat above the surface value (about
$.0095$). The Pile-CC shift fits either cohort-to-cohort variation of the kind in
App.~\ref{app:ladder} or a response to the added refresh, so the control
supports the reversal on FineWeb.

\paragraph{Weight decay and mixtures.}
Varying weight decay left the FineWeb bracket unchanged. On all three
FineWeb/GitHub mixtures $\tauwin$ was positive at every tested learning
rate.

\paragraph{Published Mousse internals.}
We restored the three differences of App.~\ref{app:impl} jointly: Lion for
non-matrix parameters, additive $10^{-5}I$ damping in place of the eigenvalue shift, and the
unadjusted matrix learning rate. The Mousse paper leaves the Lion
hyperparameters unspecified, so we used one third of the AdamW rate ($10^{-4}$), betas
$(0.9, 0.99)$ and weight decay $0.01$. Batch size, horizon and learning-rate
schedule remained ours, so the experiment isolates optimizer internals
within our training setup. The three differences were restored together,
so any outcome applies to them jointly. Our implementation at $.012$ and $.024$ ran on the same three
seeds as a control and reproduced its bracket ($-.0076$ and $+.0191$).
Under the published internals, the medians from $.008$ to $.032$ were
$-.0275$, $-.0177$, $-.0112$, $-.0040$ and $+.0034$, but the seeds at
$.032$ disagreed ($+.0034$, $+.0035$, $-.0067$). Two post-hoc extensions on the same seeds added $.048$ and $.064$, then
$.036$ and $.040$. At $.048$ and $.064$ both
cells favored earlier placement ($+.084$ and $+.27$) but lay $0.17$ to $1.1$
NLL above the best endpoint, beyond the $0.15$ stability margin. At $.036$
and $.040$ both favored earlier placement inside the margin ($+.010$ and
$+.022$), which closes a seed-consistent bracket between $.016$ and $.036$.
By Eq.~\ref{eq:crossing} the crossing is at $.032$ (seed means $-.0029$ at
$.024$ and $+.0001$ at $.032$), against $.0166$ for our implementation.
The same margin flagged two of the six runs of our implementation at $.024$,
so under both implementations the early-better side of the reversal lies
near the stability margin.

\paragraph{Other optimizers.}
Shampoo was uncompetitive in all 16 of its runs, and a grafted-Shampoo
screen found one usable learning rate of the three required on both
corpora. For a capture-coupled Muon variant, the first experiment was
inconclusive. In the second, its medians favored later placement at every tested rate ($-.0098$ to $-.0513$
over $.016$ to $.024$), while Mousse on the same setup favored earlier
placement at $.024$ ($+.0186$). These runs leave the behaviour of other optimizers open.

\section{Additional Results for Section~\ref{sec:transfer}}
\label{app:ladder}

\begin{table}[h]
\centering\footnotesize
\setlength{\tabcolsep}{4.5pt}\renewcommand{\arraystretch}{1.12}
\caption{Per-scale crossings. Ranges resample seeds within each cohort.
$^{\dagger}$A second 350M cohort. $^{\ddagger}$Independent cohorts with four fresh seeds on local three-point grids, kept separate from the main
estimates.}
\label{tab:ladderci}
\begin{tabular}{@{}lrrrc@{}}
\toprule
scale & params & \lrx & resampling range & seeds \\
\midrule
64M & 64.1M & $.029393$ & $[.029072, .029780]$ & 6 \\
124M & 124.5M & $.024049$ & $[.023101, .024160]$ & 2 \\
350M & 354.9M & $.016560$ & --- & 3 \\
350M$^{\dagger}$ & & $.016433$ & $[.016107, .016713]$ & 2 \\
695M & 695.4M & $.014327$ & $[.013943, .014757]$ & 2 \\
929M & 928.8M & $.013196$ & $[.013053, .013357]$ & 2 \\
\midrule
124M$^{\ddagger}$ & & $.026302$ & $[.025264, .026957]$ & 4 \\
695M$^{\ddagger}$ & & $.013507$ & $[.013341, .013789]$ & 4 \\
929M$^{\ddagger}$ & & $.013890$ & $[.013560, .014454]$ & 4 \\
\bottomrule
\end{tabular}

\end{table}

\paragraph{Per-scale estimates.}
Table~\ref{tab:ladderci} lists the crossing at each scale. At 124M, 695M
and 929M the main estimates rest on two seeds, so their ranges, and the
slope range $[-0.311,-0.290]$ built from them, describe only the spread of the observed seeds. Using the second 350M
cohort in place of the first changes the slope from $-0.3027$ to
$-0.3031$. Leaving one scale out
of the fit gives slopes from $-0.3103$ to $-0.2959$, with held-out
log-residuals of $+.0237$ (64M), $+.0135$ (124M), $-.0594$ (350M),
$+.0184$ (695M) and $+.0327$ (929M).

\paragraph{Independent cohorts.}
We measured 124M, 695M and 929M again with four
fresh seeds each (72 runs). All three reproduced the sign structure of the
reversal, but each cohort's interval excluded the main estimate, in both
directions: $+9.4\%$ at 124M, $-5.7\%$ at 695M and $+5.3\%$ at 929M. At
124M the difference is visible in the raw contrast: at $.024$ the main
estimate was a tie and the cohort median was $-.0067$. Variation between cohorts at a fixed scale exceeded the seed variation
within a cohort, so the per-scale ranges understate the uncertainty in the trend.
The direction of the trend is unaffected. A fit through the three cohorts
alone has slope $-0.342$ ($R^2=.971$), against $-0.300$ for the main
estimates at the same scales, and substituting it would move $\Delta_b$
from $-0.218$ to $-0.254$.

\paragraph{Held-out test at 64M.}
We fitted the four larger scales and set a tolerance band of half-width
$\exp(2\max|\text{residual}|)$ around the extrapolated 64M boundary before
running 64M. The band's half-width ($0.087$ in log space) is half the
distance to the zero-slope benchmark ($0.177$), so the test could
distinguish the two. An earlier test of the same trend at 929M, using the same band rule, had a band $3.4$ times the predicted move, wide
enough to contain the benchmark. The 64M residual ($0.024$) was smaller
than the largest residual of the four-point fit ($0.044$), and the measured
boundary lies about eight times closer in log distance to the prediction
than to the benchmark. Extrapolating the cohort-only fit instead lands
$0.106$ from the measurement, outside the band. For comparison, a
conventional log--log regression on the four scales (two residual degrees
of freedom, residual standard error $0.036$) gives a 95\% prediction
half-width of $0.254$, which contains the benchmark. The 64M experiment
used 48 matched (learning rate, seed) pairs on an 8-point grid in which every run finished and every cell was seed-consistent. Its crossing lies near the
stability edge: the rate above it ($.033$) had excess NLL of $.104$, 70\%
of the $0.15$ margin, and $.038$ and $.048$ exceeded the margin.

\paragraph{Boundary and stability edge.}
Taking the lowest unstable learning rate above the uniform-schedule optimum
as the stability edge gives $.0368$ (124M), $.0286$ (350M) and $.0245$
(929M). The ratio of boundary to edge falls from $0.65$ to $0.58$ to $0.54$,
and the edge has slope about $-0.20$ against $-0.31$ for the boundary, so the boundary falls faster than the stability edge. With three points,
partial coupling at the largest scales remains possible.

\paragraph{Boundary and optimal learning rate.}
For \lropt we used the uniform schedule $(1,188,374,561)$ at four scales,
with eight learning rates spaced by a factor of 1.35 and three fresh seeds
(96 runs), and located the optimum by a log-parabola fit around the best
grid point. Every optimum fell inside its grid. The test statistic was
$\Delta_b=\mathrm{slope}(\log\lrx)-\mathrm{slope}(\log\lropt)$ over the same
scales, with an equivalence band of $\pm0.075$, a quarter of the boundary
slope. We found $\mathrm{slope}(\log\lropt)=-0.088$ and $\Delta_b=-0.218$
(seed-bootstrap 95\% interval $[-0.261,-0.203]$), entirely outside the band.
The ratio $\lrx/\lropt$ falls $2.25\to1.96\to1.50\to1.27$ from 64M to 929M.

\paragraph{Training horizon.}
At 350M we scaled the refresh windows in proportion to the horizon. At
twice the horizon (1{,}182 steps) the FineWeb bracket was unchanged. At four
times (2{,}364 steps, three seeds) the bracket remained, with medians
$+.0018$ at $.016$ and $+.0144$ at $.024$, although one seed at $.016$
($+.0025$) exceeded the tolerance. At eight times (4{,}728 steps, three
seeds) every cell from $.008$ to $.024$ was negative or within tolerance,
with the largest median magnitude $.0041$. The $.024$ median ($-.0029$)
lay beyond the tolerance, but that cell was seed-inconsistent. The data fit either a crossing above the grid or a flattened contrast. At 64M the contrast weakened faster. At twice the horizon the crossing moved from $.0294$ to
$.0258$, at four times every cell was within tolerance except at the
stability edge ($+.033$ at $.048$), and at eight times the largest medians
were small negatives ($-.0020$ and $-.0022$).

\paragraph{Model size and tokens per parameter.}
Along the ladder, tokens per parameter fall as model size grows. We trained 64M for 107 steps and 124M for 207 steps, matching the
$0.109$ tokens per parameter of 350M at 591 steps (three seeds, full grids),
and trained 64M at two, four and eight times its horizon (two seeds). Because the contrast itself changed with training length, the planned regressions were dropped. At $0.109$ tokens per parameter, 64M and 124M
favored earlier placement at every tested rate (medians $+.002$ to
$+.033$), whereas 350M crossed at the same value. At $1.21$ tokens per
parameter 64M crossed at $.0258$, and from $2.4$ tokens per parameter the 64M contrast had largely
disappeared. At 64M the contrast weakened with training length, model size affected the
contrast beyond tokens per parameter, and the two effects remain
entangled.

\section{Additional Results for Section~\ref{sec:staleness}}
\label{app:mechanism}

\paragraph{Reference runs.}
For each (corpus, learning rate, seed) cell of the surface, a separate
reference run refreshed at 20 probe steps and recorded the spectral-width
profile (60 runs). The profiles agreed closely across seeds: their
discrepancy was $0.0105$ of the profile range. Half of each profile's total
motion was complete by steps 23 to 84, and on average $77\%$ of it occurred
before step 211 and $83\%$ before step 301.

\paragraph{Score decomposition.}
$\Delta I_w$ was positive in all 30 cells and fell from $.004$ to $.024$ in
all six corpora. $D_1$ also fell in all six (strictly monotonically in
five) and accounted for $99\%$ to $212\%$ of the net contrast. $R_w$ was
negative in 29 of the 30 cells, with median $|R_w|/D_1=0.15$. The measured
$\tauwin$ rose over the same range in five corpora.

\paragraph{Pre-specified rank test.}
The prediction compared, within each corpus, the rank order of $\Delta I_w$
across learning rates with that of the measured \tauwin. The targets and
the threshold were fixed before the reference runs. The prediction would be
withdrawn if the mean within-corpus Spearman correlation fell below $+0.40$
(a pre-specified baseline reached $+0.267$). The
observed value was $-0.667$.

\paragraph{Other staleness accounts.}
Table~\ref{tab:refutations} lists four earlier versions of the staleness
idea and their outcomes. The single-interval drift statistic reached
$+0.62$ over all 30 cells but ran opposite to the measured contrast within
the two crossing corpora, and it was measured on steps 1 to 61, before the
schedules differ.

\begin{table}[t]
\centering\footnotesize
\caption{Earlier attempts to predict the boundary from curvature staleness.}
\label{tab:refutations}
\begin{tabular}{@{}>{\raggedright\arraybackslash}p{0.455\columnwidth}>{\raggedright\arraybackslash}p{0.445\columnwidth}@{}}
\toprule
prediction & outcome \\
\midrule
Single-interval drift tracks $\tauwin$ across the surface &
runs opposite within the crossing corpora (FineWeb $-0.80$, Pile-CC
$-0.60$) \\
\addlinespace[3pt]
Refresh content (data freshness) acts as a gain &
refuted by an erasure test \\
\addlinespace[3pt]
$\tauwin$ reduces to $A\,x\ln x$ in factor age &
fails on held-out cells \\
\addlinespace[3pt]
$\tauwin$ is a generalization effect &
refuted: training and evaluation contrasts agree (corr $0.88$) \\
\bottomrule
\end{tabular}

\end{table}

\paragraph{Instrumentation limit.}
The runs stored only the spectral-width profile,
so measuring the rotation between cached and fresh bases requires storing
the cached eigenvectors.

\section{Refresh Control in Existing Implementations}
\label{app:survey}

The systems in Table~\ref{tab:survey} expose the refresh interval as a cost
control, and recent adaptive methods adjust it online
\citep{eschenhagen2025purifying,nam2026foam}. Every system we found controls the interval or its trigger, while this
work varies placement at a fixed budget. Interval names follow the cited papers
or reference implementations.

\begin{table}[t]
\centering\footnotesize
\renewcommand{\arraystretch}{1.15}
\caption{Refresh control across Kronecker-preconditioned systems.}
\label{tab:survey}
\begin{tabular}{@{}p{0.27\columnwidth}p{0.30\columnwidth}p{0.30\columnwidth}@{}}
\toprule
system & what is refreshed & exposed timing control \\
\midrule
K-FAC \citep{martens2015kfac} & factored curvature inverses & periodic
recomputation interval \\
Shampoo \citep{gupta2018shampoo,anil2020scalable} & preconditioner
roots & preconditioner update interval \\
SOAP \citep{vyas2025soap} & eigenbasis of the preconditioner &
preconditioning frequency \\
EKFAC \citep{george2018ekfac} & diagonal scalings inside a fixed
eigenbasis & rebuild interval for the eigenbasis \\
adaptive-frequency methods
\citep{eschenhagen2025purifying,nam2026foam} & eigenbasis or roots, on a
staleness trigger & adaptive update frequency (budget varies) \\
Muon \citep{liu2025muon} & (no curvature factorization) & --- \\
Mousse \citep{zhang2026mousse} & eigensystem of factored curvature EMAs &
uniform update interval $T$ ($T{=}10$ published) \\
Distributed Shampoo \citep{shi2023shampoo} & preconditioner roots &
update interval $+$ \texttt{start\_precond\-itioning\_step} warm-start \\
this work & the cached eigensystem under the implemented Mousse variant &
fixed-budget non-uniform placement \\
\bottomrule
\end{tabular}

\end{table}

\end{document}